\documentclass{article}

\usepackage{PRIMEarxiv}

\usepackage[utf8]{inputenc} 
\usepackage[T1]{fontenc}    
\usepackage{hyperref}       
\usepackage{url}            
\usepackage{booktabs}       
\usepackage{amsfonts}       
\usepackage{nicefrac}       
\usepackage{microtype}      
\usepackage{lipsum}
\usepackage{fancyhdr}       
\usepackage{graphicx}       
\usepackage{amsmath,amssymb,amsfonts}
\usepackage{enumitem}
\usepackage{booktabs} 
\usepackage{tabularx} 
\usepackage{float}
\usepackage{lineno}
\usepackage{longtable}
\title{A Scalable k-Medoids Clustering via Whale Optimization Algorithm
}

\author{
  Huang Chenan, Narumasa Tsutsumida$^{*}$ \\
  Graduate school of Science \& Engineering \\
  Saitama University \\
  \texttt{rsnaru.jp@gmail.com} \\
}

\begin{document}
\maketitle

\begin{abstract}
Unsupervised clustering has emerged as a critical tool for uncovering hidden patterns in vast, unlabeled datasets. However, traditional methods, such as Partitioning Around Medoids (PAM), struggle with scalability owing to their quadratic computational complexity. To address this limitation, we introduce WOA-kMedoids, a novel unsupervised clustering method that incorporates the Whale Optimization Algorithm (WOA), a nature-inspired metaheuristic inspired by the hunting strategies of humpback whales. By optimizing the centroid selection, WOA-kMedoids reduces the computational complexity from quadratic to near-linear with respect to the number of observations, enabling scalability to large datasets while maintaining high clustering accuracy. We evaluated WOA-kMedoids using 25 diverse time-series datasets from the UCR archive. Our empirical results show that WOA-kMedoids achieved a clustering performance comparable to PAM, with an average Rand Index (RI) of 0.731 compared to PAM's 0.739, outperforming PAM on 12 out of 25 datasets. While exhibiting a slightly higher runtime than PAM on small datasets ($<300$ observations), WOA-kMedoids outperformed PAM on larger datasets, with an average speedup of 1.7× and a maximum of 2.3×. The scalability of WOA-kMedoids, combined with its high accuracy, makes them a promising choice for unsupervised clustering in big data applications. This method has implications for efficient knowledge discovery in massive unlabeled datasets, particularly where traditional k-medoids methods are computationally infeasible, including IoT anomaly detection, biomedical signal analysis, and customer behavior clustering.

\end{abstract}

\section{Introduction}
\label{sec:introduction}
Unsupervised clustering is a crucial technique in data science that facilitates the exploration of inherent groupings within unlabeled datasets, revealing their hidden patterns and structures \cite{sinaga2020unsupervised}. This method is widely used in multiple industries such as market segmentation \cite{saunders1980cluster}, social network analysis \cite{fortunato2010community}, picture identification \cite{caron2018deep}, and gene expression data classification \cite{golub1999molecular}. It helps researchers and analysts uncover and utilize hidden relationships within the data.

Among the various approaches to unsupervised clustering, several key categories stand out, including hierarchical, density-based, and partition-based clustering \cite{xu2015comprehensive}. Hierarchical clustering constructs a tree of clusters using either a bottom-up or top-down approach, effectively capturing nested structures within data \cite{nielsen2016hierarchical}. Density-based clustering, such as DBSCAN \cite{ester1996density}, identifies clusters based on regions of high data point density, making them robust to outliers and suitable for clusters with arbitrary shapes \cite{kriegel2011density}.

Partition-based clustering algorithms are particularly noteworthy for their simplicity and efficiency \cite{yin2024rapid}. These clustering algorithms utilize distance calculations between data points to partition datasets into multiple clusters, thereby enhancing homogeneity within clusters to achieve optimal clustering outcomes. The prominent algorithms are k-means \cite{hartigan1979algorithm} and Partitioning Around Medoids (PAM) \cite{kaufman2009finding}. Although k-means efficiently excels in processing large-scale data \cite{ikotun2023k}, it struggles with datasets containing outliers or non-spherical clusters \cite{han2022data}.

By contrast, PAM offers greater flexibility and stability in clustering analysis. Unlike k-means, which uses cluster means as centroids, PAM selects actual data points as the cluster centers. This approach makes PAM more resistant to outliers and compatible with any distance measurement method. Commonly known as k-medoids, PAM represents a specific implementation of the k-medoids clustering concept \cite{kaufman2009finding}.

However, the computational complexity of PAM is $O(kn^2)$, which presents challenges in terms of its feasibility for large datasets. While CLARA \cite{kaufman2009finding} and CLARANS \cite{ng2002clarans} have been proposed to enhance the computational efficiency of PAM, they rely on sampling or randomized searches, which may lead to suboptimal medoid selection and degraded clustering quality, particularly on complex or high-dimensional datasets \cite{schubert2019faster}. In response, FastPAM \cite{schubert2019faster} was later introduced as an enhancement, achieving the same clustering accuracy as PAM while theoretically improving the speed by only $O(k)$. Thus, there remains a need for novel k-medoids algorithms that can maintain PAM's clustering accuracy while reducing its computational burden.

To address the high computational costs associated with PAM clustering, this paper introduces the application of metaheuristic algorithms as a potential solution. These algorithms leverage both global and local search capabilities to efficiently identify near-optimal solutions while reducing computational overhead, and have been widely adopted across diverse application domains \cite{tomar2024metaheuristic, elshabrawy2025review}.

Metaheuristics fall into three main categories: evolution-based, physics-based, and swarm-based methods \cite{dhiman2018emperor}. 

Evolution-based methods draw inspiration from natural selection and include Differential Evolution (DE) \cite{storn1997differential}, Genetic Algorithm (GA) \cite{mitchell1998introduction} and Covariance Matrix Adaptation Evolution Strategy (CMA-ES) \cite{hansen2001completely}. These methods have proven valuable in clustering problems, with DE notably demonstrating success in optimizing cluster centroids and improving clustering accuracy across both supervised and unsupervised tasks \cite{ali2023differential}.

Physics-based methods replicate physical phenomena. Examples include Simulated Annealing (SA) \cite{kirkpatrick1983optimization} and Gravitational Search Algorithm (GSA) \cite{rashedi2009gsa}. SA helps avoid local optima during cluster initialization \cite{selim1991simulated}, whereas GSA leverages gravitational interaction models to enhance the cluster structure \cite{hatamlou2012combined}.

Swarm-based methods emulate the collective behavior of biological swarms, featuring algorithms include Particle Swarm Optimization (PSO) \cite{kennedy1995particle}, Artificial Bee Colony (ABC) \cite{karaboga2005idea}, and Whale Optimization Algorithm (WOA) \cite{mirjalili2016whale}. These approaches have been widely used in clustering tasks, owing to their robust global search capabilities and adaptable mechanisms \cite{merwe2003data, karaboga2011novel}.

Among these three categories, swarm-based methods are particularly well suited for unsupervised clustering. Unlike evolution-based algorithms, which often require extensive parameter tuning and high computational costs, and physics-based algorithms, which often face unstable convergence, swarm-based methods strike an optimal balance between the global search capability, convergence efficiency, and implementation simplicity \cite{wang2022summary, gharehchopogh2019comprehensive}.

In particular, WOA, inspired by the bubble-net hunting behavior of humpback whales, exhibits several notable characteristics for unsupervised clustering tasks. WOA uses fewer hyperparameters than other swarm-based approaches and achieves a more effective balance between exploration and exploitation through its spiral update mechanism \cite{mirjalili2016whale}. This balance has contributed to its successful application in various optimization problems, including clustering \cite{gharehchopogh2019comprehensive}. While Nasiri and Khiyabani successfully integrated WOA with the k-means algorithm to improve clustering performance \cite{nasiri2018whale}, the application of WOA to k-medoids clustering has, to the best of our knowledge, not yet been explored, presenting a clear research gap that this study aims to address.

Previous heuristic approaches to k-medoids clustering, such as GA-based k-medoids \cite{sheng2006genetic} and PSO combined with SA for k-medoids \cite{zhou2013k}, have relied on complex solution encoding and extensive parameter tuning. In contrast, WOA offers a structurally simpler optimization framework with fewer control parameters. Additionally, while these earlier algorithms were designed for continuous optimization tasks, WOA can be effectively adapted to discrete problems, such as the medoid selection task in k-medoids clustering, through simple index mapping and rounding techniques. These features position the WOA as a promising candidate for enhancing the performance of k-medoids clustering.

This study introduces WOA-kMedoids, a novel approach designed to enhance the efficiency of k-medoids clustering. Specifically, the proposed method employs the WOA to optimize medoid selection, with the expectation of substantially reducing computational costs compared to PAM while maintaining clustering accuracy.

The main contributions of this study are as follows:
\begin{itemize}
\item We present WOA-kMedoids, a novel hybrid method that integrates the Whale Optimization Algorithm with k-medoids clustering.
\item We enhance traditional k-medoids scalability through WOA-based efficient medoid selection.
\item Our approach achieves linear computational time relative to data size, enabling big data clustering scenarios.
\item Through extensive experiments on multiple benchmark datasets, we demonstrate that WOA-kMedoids maintains PAM's accuracy level while improving computational efficiency.
\end{itemize}

The remainder of this paper is organized as follows.  
Section~\ref{sec:related} reviews the foundational PAM and WOA algorithms that form the basis of our approach.  
Section~\ref{sec:woa-k} presents the WOA-kMedoids method and outlines its algorithmic design and computational characteristics.  
Section~\ref{sec:experiments} describes our experimental methodology, including dataset selection, distance metrics, and evaluation criteria.  
Section~\ref{sec:results} presents and analyzes our results by comparing WOA-kMedoids with established baseline methods.  
Section~\ref{sec:discussion} discusses our key findings, addresses limitations, and explores future directions.  
Section~\ref{sec:conclusion} concludes the paper with a summary and perspectives for future work.

\section{Related Work}
\label{sec:related}

\subsection{Partitioning Around Medoids}
The PAM algorithm is an implementation of k-medoids clustering \cite{kaufman2009finding}. It identifies the most representative data points in a dataset to serve as centroids of the clusters. In PAM, two phases called BUILD and SWAP are utilized to group data points in a way that minimizes the total distance within each cluster based on a similarity measurement technique such as Euclidean distance. The BUILD phase involves selecting the initial medoids, whereas the SWAP phase is responsible for updating the medoids.

\subsubsection{BUILD phase}
\begin{enumerate}
    \item Calculate the total distance from each data point to all others. The point with the minimum total distance is selected as the first centroid.
    \item Choose subsequent centroids that minimize the total distance cost. The total distance cost is the sum of the distances from each point to its nearest centroid.
    \item Repeat step 2 until $k$ centroids are selected.
\end{enumerate}

The BUILD phase iteratively selects $k$ initial medoids. For each selection, all the data points are evaluated as potential medoids. The time complexity is $O(kn^2)$, where $n$ is the number of data points, and $k$ is the number of medoids.

\subsubsection{SWAP phase}
\begin{enumerate}
    \item Evaluate replacing each centroid with a non-centroid point. A greedy approach is used to determine the swap that minimizes the total distance cost. Execute the swap if the total distance cost is reduced.
    \item Repeat step 1 for all centroids to optimize the centroid combination.
\end{enumerate}

The computational complexity of the SWAP phase is $O((n-k)^2)$.

The total computational cost of PAM is high when dealing with large datasets. Hence, there is a need for clustering methods that improve the computational efficiency.

\subsection{Whale Optimization Algorithm}
The WOA is an optimization algorithm that emulates the hunting behavior of humpback whales \cite{mirjalili2016whale}. This is inspired by the "bubble-net feeding" behavior observed in humpback whales.

\subsubsection{Bubble-net feeding}
A whale uses a bubble-net feeding strategy by creating bubbles in a spiral path from the seabed to the surface, which is used to encircle and capture schools of fish as prey. The WOA simulates this behavior through \textit{Encircling prey} and \textit{Spiral upward movement}.

\paragraph{Encircling prey}
Let the maximum number of iterations be denoted as $t_{max}$, and among $L$ whales, the position of  the prey at the current time $t$ is represented by $\vec{X^\ast}(t)$. The whale adjusts its position relative to $\vec{X^\ast}(t)$ based on Equation \ref{equ_1}:
\begin{equation} \label{equ_1}
\left\{
\begin{aligned}
    & \vec{X}(t+1) = \vec{X^\ast}(t) - \vec{A} \cdot \vec{D} \\
    & \vec{D} = \lvert \vec{C} \cdot \vec{X^\ast}(t) - \vec{X}(t) \rvert ,
\end{aligned}
\right.
\end{equation}

where $\vec{X}(t)$ denotes the position of a certain whale at time $t$ and $\vec{D}$ is the absolute distance between the whale and the prey's position. $\vec{A}$ and $\vec{C}$ are coefficient vectors defined by equation 2:
\begin{equation} \label{equ_2}
\left\{
\begin{aligned}
    \vec{A} &= 2 \vec{a} \cdot \vec{r} - \vec{a} \\
    \vec{a} &= 2 - 2 (t / t_{max}) \\
    \vec{C} &= 2 \cdot \vec{r},
\end{aligned}
\right.
\end{equation}

where $\vec{r}$ is a random vector within the range of [0,1] and $\vec{a}$ linearly decreases from two to zero. The range of $\vec{A}$ becomes [-2, 2].

In the WOA, encircling prey behavior is only performed when $\vec{A}$ is within [-1, 1]. This corresponds to ecological observations that whales engage in hunting behavior only when they are in close proximity to their prey. Furthermore, employing a random vector $\vec{r}$ enables the exploration of any position between the whale and prey within the search space.

\paragraph{Spiral upward movement}
To simulate the spiral upward movement of whales during prey capture, Equation \ref{equ_3} is used:
\begin{equation} \label{equ_3}
\left\{
\begin{aligned}
    & \vec{X}(t+1) = \vec{D\prime} \cdot e^{bl} \cdot cos(2 \pi l) + \vec{X^\ast}(t) \\
    & \vec{D\prime} = \lvert \vec{X^\ast}(t) - \vec{X}(t) \rvert ,
\end{aligned}
\right.
\end{equation}

where $\vec{D\prime}$ is the absolute distance between the whale and the prey,
$b$ is a constant that defines the shape of the logarithmic spiral and $l$ is randomly chosen from the range $[-1, 1]$.

Notably, when whales hunt using the bubble-net feeding strategy, they encircle the prey while moving in a spiral, demonstrating both of the aforementioned behaviors. Hence, bubble-net feeding is performed based on random probability $p$ as shown in Equation \ref{equ_4}:
\begin{equation} \label{equ_4}
\left\{
\begin{aligned}
    &\vec{X}(t+1) = \vec{X^\ast}(t) - \vec{A} \cdot \vec{D} \quad (p < 0.5) \\
    &\vec{X}(t+1) = \vec{D\prime} \cdot e^{bl} \cdot cos(2 \pi l) + \vec{X^\ast}(t) \quad (p \geq 0.5)
\end{aligned}
\right.
\end{equation}

\subsubsection{Searching for Prey}
The behavior of whales is not limited to bubble-net feeding but also includes exploration activities. Consequently, within the framework of the WOA, the algorithm initiates the simulation of whales engaging in random search behavior for prey when $1 < \lvert\vec{A}\rvert < 2$. As illustrated in Equation \ref{equ_5}, whale $\vec{X}(t)$ moves away from the optimal position $\vec{X^\ast}(t)$ to explore a broader search.
\begin{equation} \label{equ_5}
\left\{
\begin{aligned}
    & \vec{X}(t+1) = \vec{X}_{rand}(t) - \vec{A} \cdot \vec{D} \\
    & \vec{D} = \lvert \vec{C} \cdot \vec{X}_{rand}(t) - \vec{X}(t) \rvert,
\end{aligned}
\right.
\end{equation}

where $\vec{X}_{rand}(t)$ denotes the current position at time $t$ of a whale chosen randomly from the pod.

The WOA showcases its capability to thoroughly investigate the target area using bubble-net feeding and to cover the entire search space by searching for prey. This capability enables the WOA to identify globally optimal solutions to optimization problems.

Within the WOA framework, it is essential to consider not only the three exploration behaviors mentioned earlier but also the critical task of prey localization. The prey's location is determined by the fitness function, denoted as $Fitness$, which varies depending on the characteristics of the optimization problem. A well-configured $Fitness$ enhances the adaptability and problem-solving abilities of the WOA in various problem domains.

\section{WOA-kMedoids}
\label{sec:woa-k}
We propose a WOA-kMedoids, which is a fast unsupervised clustering approach that introduces WOA to k-medoids.
In this approach, each whale $\vec{X}$ represents a candidate solution consisting of $k$ data-point indices selected as medoids. This approach aligns naturally with the k-medoids criterion, which requires cluster centers to be actual points from the dataset, framing medoid selection as a discrete combinatorial optimization problem.

Because the WOA was originally intended for continuous optimization, we adapt it for discrete medoid selection by incorporating a rounding operation after each whale's position update. This ensures that each whale's position vector consistently represents valid data-point indices. Furthermore, in each iteration, the WOA employs various search strategies, including encircling the current best solution, spiral searching, and random exploration, to stochastically adjust the positions of the whales. This allows the exploration of promising medoid combinations within a discrete solution space.

During the iterative optimization process, the fitness of each whale is evaluated based on the total intra-cluster distance induced by the medoid combination. The whale with the lowest intra-cluster distance is designated as the current optimal solution, $\vec{X^\ast}$, which subsequently guides the position updates of the remaining whales. Through successive iterations and interactions, the whale population converges toward an optimal or near-optimal set of medoids, thereby effectively enhancing the clustering performance.

The detailed algorithmic steps and complexity analysis are as follows.

\begin{enumerate}[label=Step \arabic*.,leftmargin=*]
\item Distance Calculation \par
Compute pairwise distances between data points using a selected distance metric.

\item WOA Parameter Configuration \par
Define the following parameters: the total number of whales $L$, the number of clusters $k$, the number of iterations $t_{max}$, and the minimum number of data points required in each cluster.

\item Whale Initialization \par
In WOA-kMedoids, the whales represent combinations of cluster centroids.
In the initial state, a set of $k$ initial centroids is randomly selected from the dataset. Therefore, in the initial state $(t=0)$, a whale $\vec{X}(0)$ is positioned at $k$ random data points as medoids $(c_1,c_2,...,c_k)$ as shown in Equation \ref{equ_6}.
\begin{equation} \label{equ_6}
    \vec{X}(0) = (c_1,c_2,...,c_k)
\end{equation}

Given that the total number of whales is $L$, there will exist $L$ sets of different initial center combinations.

\item Prey Identification  \par
Assign each data point to its nearest centroid for each whale. Determine the prey location using $Fitness$, which minimizes the total distances among data points within each cluster (Equation \ref{equ_7}).
\begin{align} \label{equ_7}
    Fitness = min\sum_{i=1}^{k} \sum_{data \in C_i} Dist(data,c_i),
\end{align}

where $C_i$ denotes the set of data points in cluster $i$ and $Dist(data,c_i)$ represents the distance between data point $data$ and center $c_i$.
Among $L$ whales, the whale with the minimum $Fitness$ is determined as prey $\vec{X^\ast}(t)$ at time $t$.

\item Whale Pod Update \par
Adjust whale positions based on bubble-net feeding and prey-searching behaviors governed by random probability $p$ (Equation \ref{equ_8}). This enhances the stochastic nature of the WOA, thereby enabling it to explore a broader area.
\newcommand{\round}[1]{\left\lfloor #1 \right\rceil} 
\begin{equation} \label{equ_8}
    \vec{X}(t+1) = \left\{\begin{aligned}
        &\round{\vec{X^\ast}(t) - \vec{A} \cdot \vec{D}} \quad (p < 0.5 \; \text{and} \; \left|\vec{A}\right| \leq 1) \\
        &\round{\vec{D\prime} \cdot e^{bl} \cdot cos(2 \pi l) + \vec{X^\ast}(t)} \quad (p \geq 0.5) \\
        &\round{\vec{X}_{rand}(t) - \vec{A} \cdot \vec{D}} \quad (p < 0.5 \; \text{and} \; \left|\vec{A}\right| > 1),
\end{aligned}\right.
\end{equation}

where the rounding function $\round{\cdot}$ is defined as the rounding to the nearest integer. This is used to confirm that the revised location corresponded to the actual data point in the dataset.

\item Centroid Identification \par
In each iteration, all $L$ whales first calculate their respective fitness values and identify prey locations through Step 4, then update their positions according to the search strategy in Step 5. This process is repeated $t_{max}$ times. After all iterations, the centroid combination corresponding to the whale with the minimum $Fitness$ value is selected as the optimal solution.

\item Clustering Finalization \par
Assign each data point to its nearest centroid from the optimal combination identified in Step 6.

\end{enumerate}

The time complexity of WOA-kMedoids for time series depends on the total number of whales $L$, number of iterations $t_{max}$, complexity of the fitness function $Fitness$, number of clusters $k$, and number of data points $n$.

Increasing $L$ and $t_{max}$ linearly contributes to the stability of the optimization. The complexity of $Fitness$ directly affects the time complexity of the WOA. A more complex $Fitness$ requires additional computational resources, resulting in a longer computation time. $Fitness$ in this study requires computing the distance from each data point to all $k$ centroids and choosing the closest center. With $n$ data points, the total time complexity for calculating the $Fitness$ is $O(kn)$. Thus, the time complexity of WOA-kMedoids for the time series is estimated to be $O(Lt_{max}kn)$. Compared with the computational complexity of PAM, which is expressed as $O(kn^2)$, the WOA has the potential to decrease the computational effort and enhance the clustering efficiency, especially when dealing with large values of $n$.

\section{Experiments}
\label{sec:experiments}
To examine whether the WOA-kMedoids preserves the same level of precision as PAM while increasing computational efficiency, we selected 25 time-series datasets from the UCR Time Series Archive \cite{UCRArchive2018} for our evaluation, using the Rand Index (RI) \cite{rand1971objective} and the execution time of clustering to assess the results.

Within the WOA-kMedoids framework, Dynamic Time Warping (DTW) \cite{sakoe1978dynamic} is employed as the primary distance metric to assess the similarity between time-series data points.

\subsection{Dynamic Time Warping}
\begin{figure}[htbp]
\centering
\includegraphics[width=\linewidth]{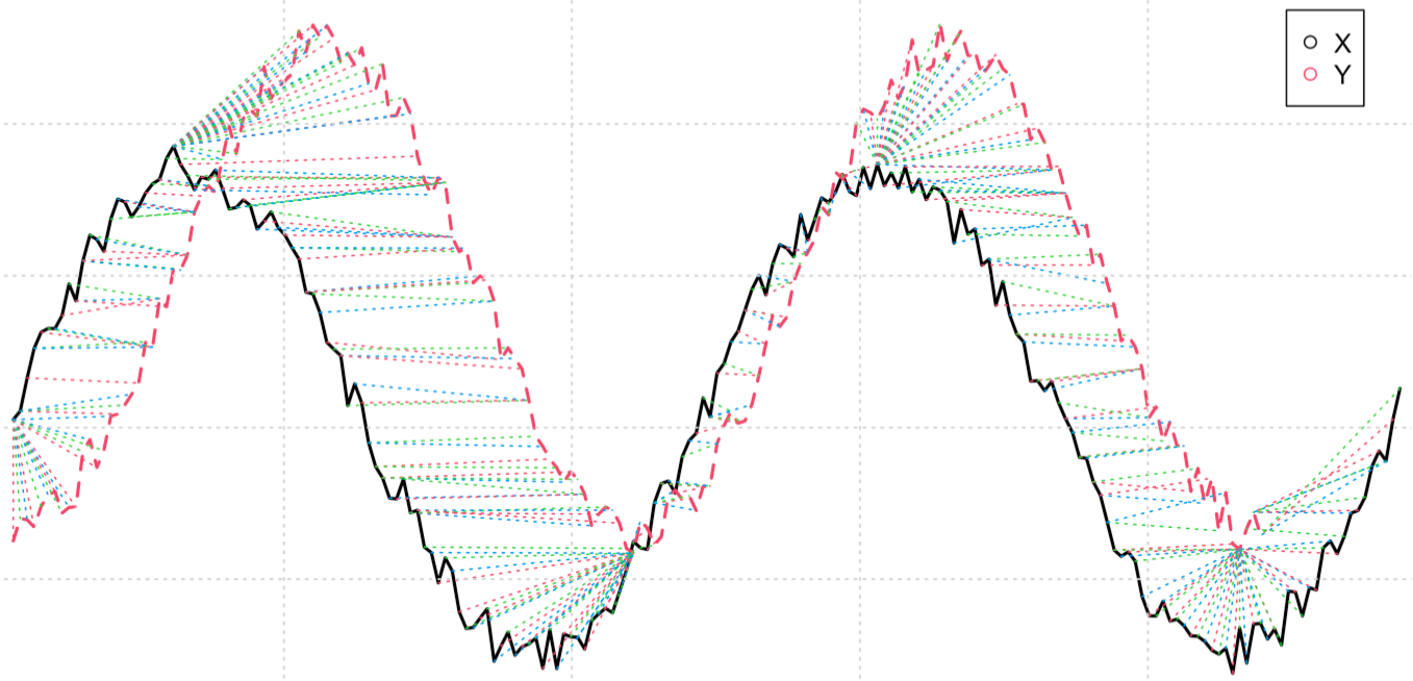}
\caption{Conceptual diagram of applying Dynamic Time Warping (DTW) to time series data $X$ and $Y$.}
\label{fig_1}
\end{figure}

DTW is a method used to assess the similarity of time-series data \cite{sakoe1978dynamic}. It enables many-to-one mapping of individual time-series data points, facilitating the assessment of similarity between time series of varying lengths. Here, $X = (x_1, x_2, \dots, x_i, \dots, x_n)$ and $Y = (y_1, y_2, \dots, y_j, \dots, y_m)$ as an example (Figure \ref{fig_1}). $n$ and $m$ represent the number of elements in time series $X$ and $Y$, respectively, which can be different integers.

As shown in Equation \ref{equ_9}, the distance $d(x_i,y_j)$ between $x_i$ and $y_j$ is defined as the element $Mat(i,j)$ within a distance matrix $Mat$ of size $n \times m$.
\begin{equation} \label{equ_9}
\left\{
\begin{aligned}
    & Mat(i,j) = d(x_i,y_j)  \\
    & d(x_i,y_j) = (x_i - y_j)^2
\end{aligned}
\right.
\end{equation}

The minimum path is calculated from the distance matrix through dynamic programming according to Equation \ref{equ_10}, which is then defined as the DTW distance.
\begin{equation} \label{equ_10}
    Mat_{DTW}(i,j) = Mat(i,j) + \min\left(\begin{aligned}
        & Mat_{DTW}(i-1,j) \\
        & Mat_{DTW}(i,j-1) \\
        & Mat_{DTW}(i-1,j-1)
    \end{aligned}\right),
\end{equation}

where $Mat_{DTW}(i,j)$ denotes the DTW distance between $x_i$ and $y_j$.
To improve the accuracy and speed of DTW, a window constraint can be implemented. The window size determines the maximum number of elements of $Y$ that can be associated with $x_i$, or conversely, the maximum number of elements of $X$ that can be associated with $y_j$. When the window size is set to $w$, the value of $x_i$ is constrained to data within the interval $[y_{j-w},y_{j+w}]$.

\subsection{UCR Time Series Archive}
The UCR Time Series Archive is a widely recognized benchmark dataset for research in time-series clustering and classification. It contains various types of time-series data, including electrocardiogram signals, images, sensor data, and synthetic data. The archive provides clean, pre-processed data with no missing values, making it immediately ready for analysis. Each dataset is pre-divided into training and test subsets by the archive. This study focuses solely on 25 test datasets extracted from the UCR Archive, as shown in Table \ref{tab01}. Each dataset contains precise classification labels, as defined by the researchers.

\begin{longtable}{@{}l>{\raggedright\arraybackslash}p{1.8cm}cccr@{}}
\caption{Summary of UCR Time Series Archive Datasets Utilized in this study. The dataset characteristics include the dataset name, data type classification, total number of instances (N), temporal sequence length (dimensionality), number of distinct clusters (k), and the optimal DTW window parameter ($w^\ast$) as specified in the original dataset.}
\label{tab01}\\
\toprule
Dataset & Type & Number & Length & \(k\) & \(w^\ast\) \\
\midrule
\endfirsthead

\multicolumn{6}{c}{\tablename\ \thetable\ -- \textit{Continued from previous page}} \\
\toprule
Dataset & Type & Number & Length & \(k\) & \(w^\ast\) \\
\midrule
\endhead

\midrule
\multicolumn{6}{r}{\textit{Continued on next page}} \\
\endfoot

\bottomrule
\endlastfoot

\footnotesize
Lightning7 & Sensor & 73 & 319 & 7 & 5 \\
Worms & Motion & 77 & 900 & 5 & 9 \\
ACSF1 & Device & 100 & 1460 & 10 & 4 \\
HouseTwenty & Device & 119 & 2000 & 2 & 33 \\
Earthquakes & Sensor & 139 & 512 & 2 & 6 \\
SmoothSubspace & Simulated & 150 & 15 & 3 & 1 \\
Fish & Image & 175 & 463 & 7 & 4 \\
PowerCons & Power & 180 & 144 & 2 & 3 \\
InsectEPGRegularTrain & EPG & 249 & 601 & 3 & 11 \\
SyntheticControl & Simulated & 300 & 60 & 6 & 6 \\
EOGVerticalSignal & EOG & 362 & 1250 & 12 & 2 \\
RefrigerationDevices & Device & 375 & 720 & 3 & 8 \\
EthanolLevel & Spectro & 500 & 1751 & 4 & 1 \\
SemgHandGenderCh2 & Spectrum & 600 & 1500 & 2 & 1 \\
CBF & Simulated & 900 & 128 & 3 & 11 \\
TwoLeadECG & ECG & 1139 & 82 & 2 & 4 \\
MoteStrain & Sensor & 1252 & 84 & 2 & 1 \\
FaceAll & Image & 1690 & 131 & 14 & 3 \\
InsectWingbeatSound & Sensor & 1980 & 256 & 11 & 1 \\
Yoga & Image & 3000 & 426 & 2 & 7 \\
UWaveGestureLibraryAll & Motion & 3582 & 945 & 8 & 4 \\
TwoPatterns & Simulated & 4000 & 128 & 4 & 4 \\
ECG5000 & ECG & 4500 & 140 & 5 & 1 \\
ElectricDevices & Device & 7711 & 96 & 7 & 14 \\
StarLightCurves & Sensor & 8236 & 1024 & 3 & 16 \\
\end{longtable}

These datasets encompass diverse sizes and sequence characteristics, providing an ideal testbed for evaluating clustering algorithms across different complexity levels. Table~\ref{tab01} presents the key dataset attributes: the "Number" column shows the total samples per dataset (ranging from 73 to 8,236), while the "Length" column indicates each sample's dimensionality at time points. The datasets contain between two and 14 clusters ($k$), and each includes a validated optimal DTW window size ($w^\ast$) specified in the Archive.

\subsection{Experimental Setup and evaluation}
WOA-kMedoids was applied to 25 test datasets from the UCR Time Series Archive with the following parameters: whale population size $L=50$, maximum iterations $t_{max} = 200$, and minimum cluster size of two data points.

We compared DTW-WOA-kMedoids (our method using DTW distance) with DTW-PAM (PAM using DTW distance) to evaluate performance differences. We also included ED-PAM (PAM using Euclidean distance) to assess the accuracy variations between DTW and Euclidean metrics. For consistency, we used the optimal DTW window size ($w^\ast$) for both the DTW-PAM and DTW-WOA-kMedoids.

With the true clustering labels in the UCR datasets, the clustering results were evaluated by RI. The RI value quantifies the degree of agreement between the results of the clustering and the true clustering labels, as defined by Equation \ref{equ_11}.
\begin{equation} \label{equ_11}
    RI = \frac{a+b}{a+b+c+d}
\end{equation}

where $a$ is the number of element pairs classified into the same cluster in both the clustering results and the true clustering labels; $b$ is the number of element pairs classified into different clusters in both the clustering results and the true labels; $c$ is the number of element pairs classified into the same cluster but into different clusters in the true labels; and $d$ is the number of element pairs classified into different clusters in the clustering results but the same cluster in the true clustering labels.

The numerator ($a+b$) represents the number of element pairs for which the clustering results and true labels agree. The denominator ($a+b+c+d$) represents the total number of possible pairs of elements. Ideally, when the clustering results perfectly match the true clustering labels, the RI value is one. Therefore, the closer the RI value is to one, the more accurate the clustering results reflect true clustering labels.

To evaluate whether the differences in clustering performance among the different methods were statistically significant, we first applied the Friedman test \cite{friedman1937use}, a non-parametric test commonly used to compare three or more results under identical experimental conditions. The null hypothesis assumes no significant difference in performance among the methods; therefore, a smaller $p$-value indicates stronger evidence that the methods perform significantly differently from each other. Following this, we employed the Conover post-hoc test \cite{conover1999practical} to identify the specific pairs of methods that exhibit statistically significant differences.

Additionally, we compared the execution time of clustering for the UCR dataset using WOA-kMedoids and PAM to evaluate the computational efficiency, excluding distance calculation. To assess the statistically significance of runtime differences, we employed the Wilcoxon signed-rank test \cite{conover1999practical}, which computes pairwise runtime differences across all 25 datasets. The test statistic $W$ represents the smaller signed-rank sum, indicating a directional bias in the performance differences. For $n$ paired observations, the maximum rank sum is given by $\frac{n(n+1)}{2}$, which serves as a reference for interpreting the deviation of the observed $W$ from a balanced distribution. This test enables us to assess whether the runtime differences across datasets exhibit a statistically significant directional trend.

Beyond the clustering accuracy and computational efficiency, we analyzed the convergence, diversity, and parameter sensitivity characteristics of WOA-kMedoids using the EOGVerticalSignal ($Number$=362, $k$=12) dataset as a representative example. 

Convergence was assessed by tracking the fitness function (Equation 7) across 200 iterations from 10 independent runs. 

To measure population diversity, we counted distinct data points selected as medoids (the unique medoids) across all 50 whales. Whereas each whale selects $k$ medoids (yielding $k \times 50$ selections in total), multiple whales often choose the same data points, creating an overlap. Therefore, the actual number of unique medoids ranges from $k$ (when all whales converge to identical solutions) to $k \times 50$ (when no overlap exists between whales' selections). Higher unique medoids indicate greater population diversity, which reflects the exploration capability of the algorithm.

For the parameter sensitivity analysis, the population size $L$ was varied across $\{10, 30, 50, 70, 90\}$ and the maximum iteration count $t_{\text{max}}$ across $\{20, 50, 100, 150, 200, 250, 300\}$, yielding 35 parameter combinations. Each configuration was evaluated based on clustering accuracy (RI value) and convergence quality (fitness value) to identify the optimal parameter settings.

All analyses were conducted in an R environment (Version 4.3.2) on a Mac mini (M2) equipped with a 3.49 GHz CPU and 16 GB of memory. All implementations used pure R code without C++ integration, to ensure fair computational time comparisons.

\section{Results}
\label{sec:results}

\subsection{Clustering Accuracy Comparison}
The RI values of the ED-PAM, DTW-PAM, and DTW-WOA-kMedoids for the 25 UCR time series datasets are presented in Table \ref{tab02}. The average RI values for DTW-WOA-kMedoids and DTW-PAM were closer (0.731 and 0.734, respectively), while that for ED-PAM (0.722) was lower than others. In 80\% of the datasets, the RI values of DTW-WOA-kMedoids surpassed that of ED-PAM, with the RI values being consistent in 16\% of the datasets. Furthermore, DTW-WOA-kMedoids outperformed DTW-PAM on 36\% of the datasets, with identical RI values observed in 20\% of the cases.

\begin{longtable}{@{}lccc@{}}
\caption{RI values for PAM with Euclidean Distance (ED-PAM), PAM with Dynamic Time Warping (DTW-PAM), and our approach (DTW-WOA-kMedoids).}
\label{tab02}\\
\toprule
Dataset & ED-PAM & DTW-PAM & DTW-WOA-kMedoids \\
\midrule
\endfirsthead

\multicolumn{4}{c}{\tablename\ \thetable\ -- \textit{Continued from previous page}} \\
\toprule
Dataset & ED-PAM & DTW-PAM & DTW-WOA-kMedoids \\
\midrule
\endhead

\midrule
\multicolumn{4}{r}{\textit{Continued on next page}} \\
\endfoot

\bottomrule
\endlastfoot

Lightning7 & 0.803 & 0.807 & \textbf{0.811} \\
Worms & 0.647 & \textbf{0.667} & 0.656 \\
ACSF1 & 0.721 & 0.697 & \textbf{0.732} \\
HouseTwenty & 0.567 & \textbf{0.628} & \textbf{0.628} \\
Earthquakes & 0.500 & 0.502 & \textbf{0.532} \\
SmoothSubspace & 0.729 & \textbf{0.823} & 0.771 \\
Fish & 0.768 & 0.771 & \textbf{0.792} \\
PowerCons & \textbf{0.846} & \textbf{0.846} & \textbf{0.846} \\
InsectEPGRegularTrain & \textbf{1} & \textbf{1} & \textbf{1} \\
SyntheticControl & 0.819 & 0.880 & \textbf{0.887} \\
EOGVerticalSignal & \textbf{0.856} & \textbf{0.856} & \textbf{0.856} \\
RefrigerationDevices & 0.540 & \textbf{0.600} & 0.579 \\
EthanolLevel & 0.612 & 0.621 & \textbf{0.624} \\
SemgHandGenderCh2 & \textbf{0.511} & \textbf{0.511} & 0.499 \\
CBF & 0.667 & \textbf{0.734} & 0.713 \\
TwoLeadECG & 0.501 & \textbf{0.585} & 0.558 \\
MoteStrain & 0.737 & \textbf{0.804} & 0.740 \\
FaceAll & 0.873 & \textbf{0.917} & 0.889 \\
InsectWingbeatSound & 0.885 & 0.885 & \textbf{0.886} \\
Yoga & \textbf{0.500} & \textbf{0.500} & \textbf{0.500} \\
UWaveGestureLibraryAll & 0.894 & \textbf{0.901} & 0.897 \\
TwoPatterns & 0.633 & 0.632 & \textbf{0.637} \\
ECG5000 & 0.673 & \textbf{0.738} & 0.683 \\
ElectricDevices & 0.738 & \textbf{0.802} & 0.784 \\
StarLightCurves & 0.765 & 0.765 & \textbf{0.768} \\
Average of all datasets & 0.711 & \textbf{0.739} & 0.731 \\
\end{longtable}

The Friedman test yielded a $\chi^2$ value of 20.425 and $p$-value of $3.671 \times 10^{-5}$ ($p < 0.001$), indicating a significant difference in the RI values among the three methods  (see Table \ref{tab03}).
\begin{table}[H]
    \centering
    \caption{Results of the Friedman test.}
    \begin{tabular}{cc}
    \toprule
    Friedman $\chi^2$ & $p$-value \\
    \midrule
    20.425 & $3.671 \times 10^{-5}$ \\
    \bottomrule
    \end{tabular}
    \label{tab03}
\end{table}

The Conover test revealed a statistically significant difference between DTW-WOA-kMedoids and ED-PAM ($p=1.4 \times 10^{-6}$), but no significant difference between DTW-WOA-kMedoids and DTW-PAM ($p=0.230$) (see Table \ref{tab04}). These results indicate that, while DTW-WOA-kMedoids and DTW-PAM perform similarly in terms of accuracy, both methods significantly outperform ED-PAM, demonstrating the effectiveness of DTW for time-series clustering.
\begin{table}[H] 
    \centering
    \caption{Result of Conover test.}
    \begin{tabular}{lcc}
    \toprule
           & ED-PAM   & DTW-PAM \\
    \midrule
    DTW-PAM & $2.4 \times 10^{-6}$ & -       \\
    WOA-DTW-kMedoids & $1.4  \times 10^{ -6}$ & 0.230    \\
    \bottomrule
    \end{tabular}
    \label{tab04}
\end{table}

\begin{figure}[htbp]
\centering
\includegraphics[width=0.8\columnwidth]{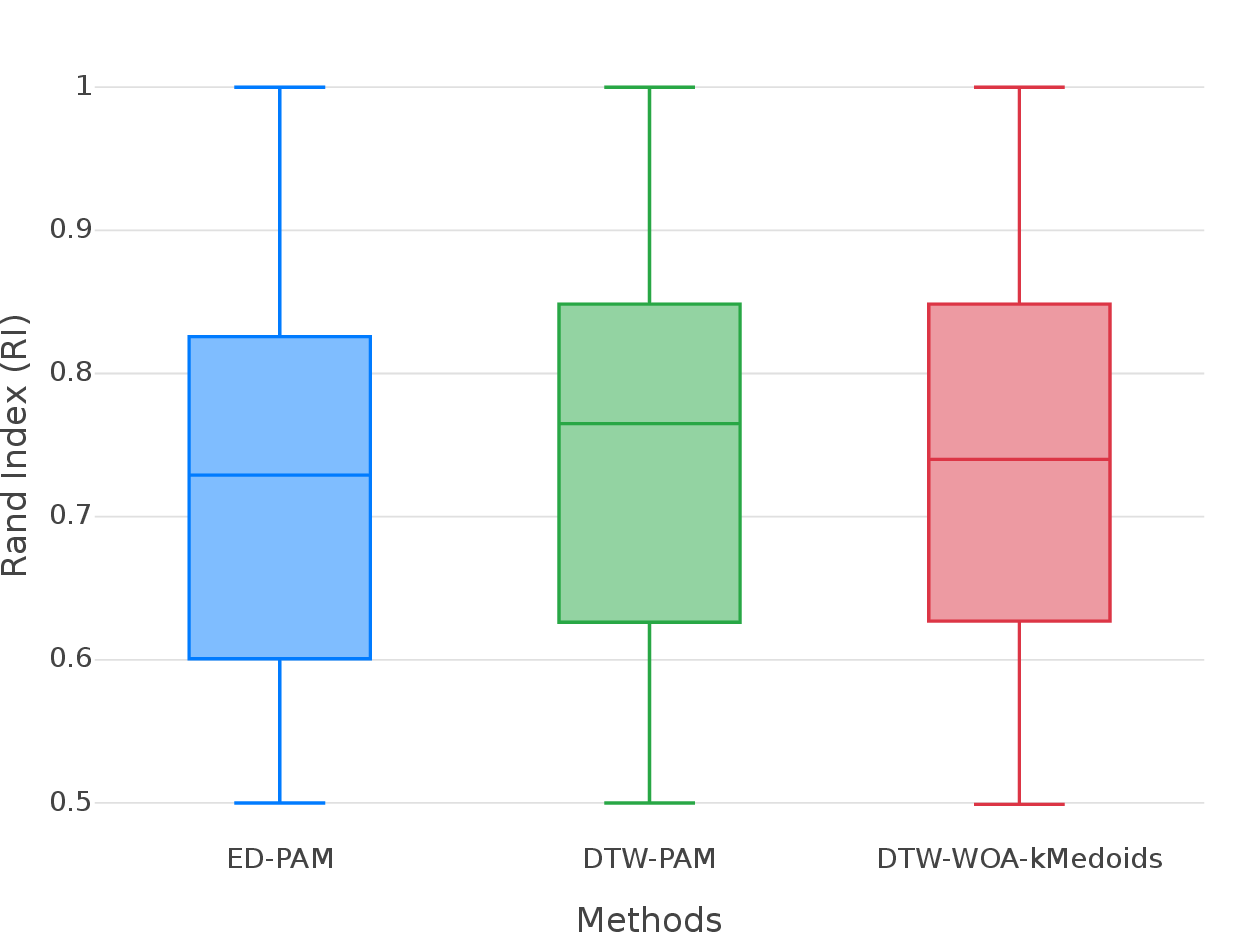}
\caption{Distribution of RI values across 25 UCR datasets for three clustering methods.}
\label{fig_2}
\end{figure}

To visualize the performance distribution across the 25 datasets, Figure \ref{fig_2} displays boxplots of the RI values for the three methods. DTW-PAM achieved the highest median (0.765), followed by DTW-WOA-kMedoids (0.740), and ED-PAM (0.729). Notably, DTW-WOA-kMedoids exhibited the lowest variance (0.019), with DTW-PAM (0.020) and ED-PAM (0.020) showing slightly higher variability. This indicates that DTW-WOA-kMedoids delivers consistent performance across heterogeneous datasets.

However, these median values should be interpreted with caution because of the heterogeneous nature of the datasets. The data encompassed diverse domains (sensor, image, and medical signals), dimensions (15 to 2000), and sample sizes (73 to 8236). This diversity suggests that median and variance differences alone may not fully reflect how each method performs across different data characteristics.

\subsection{Computational Efficiency Analysis}
\begin{figure}[htbp]
\centering
\includegraphics[width=\linewidth]{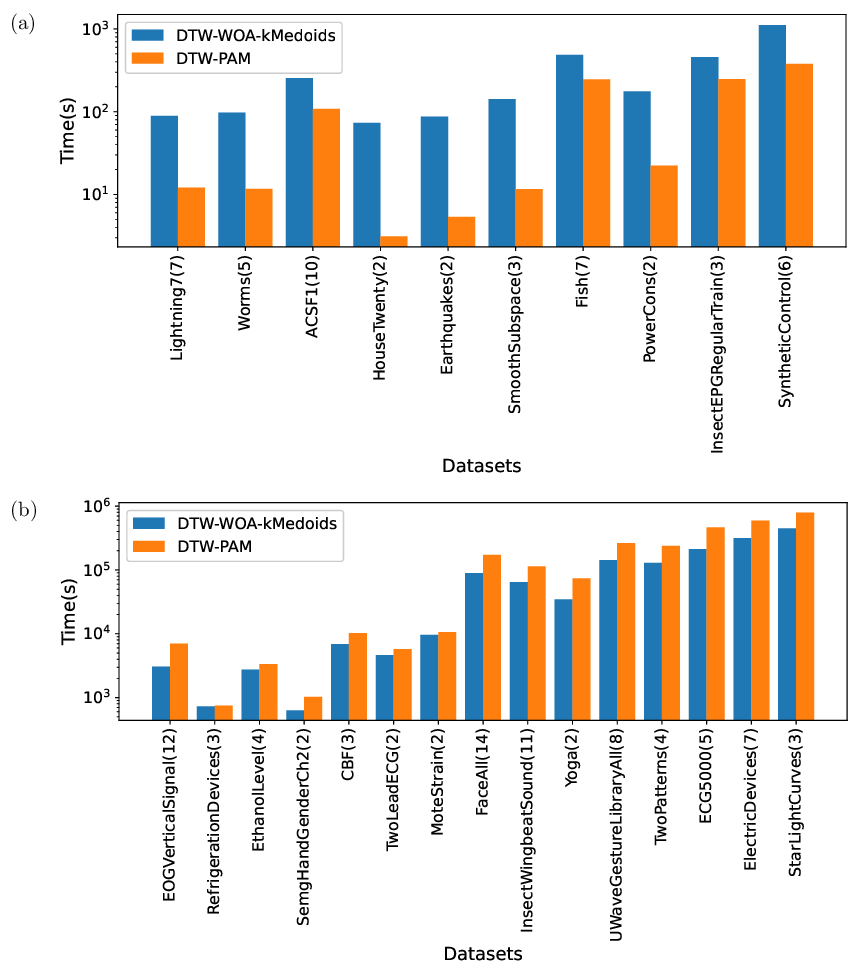}
\caption{Computational time (seconds) in DTW-WOA-kMedoids and DTW-PAM for (a) dataset having 73 to 300 data points and (b) dataset having 362 to 8236 data points. Numbers within () indicate the values of $k$. Note that the execution time for calculating DTW distance was excluded.}
\label{fig_3}
\end{figure}

The computational times of DTW-WOA-kMedoids and DTW-PAM were compared, and the results  are summarized in Figure \ref{fig_3}. Given that both approaches employed DTW, we excluded the computational time for calculating the DTW distance.

For datasets with fewer than 300 data points, the execution of DTW-WOA-kMedoids was slower than that of DTW-PAM (Figure \ref{fig_3}(a)). However, as the dataset size increased, the execution of DTW-WOA-kMedoids tended to be faster than that of DTW-PAM (Figure \ref{fig_3}(b)).

To further support this observation, Table \ref{tab05} summarizes the speedup of the WOA over PAM on datasets containing more than 300 data points. As shown in the table, WOA-kMedoids achieved an average speedup of approximately 1.7×, with the largest gain observed on the \textit{EOGVerticalSignal} dataset (approximately 2.3×).

The execution times of the two methods were influenced by both the number of data points ($Number$) and clusters ($k$). For datasets with similar numbers of data points, such as \textit{EOGVerticalSignal} and \textit{RefrigerationDevices}, the execution time differences were mainly due to the varying number of clusters. Conversely, when comparing datasets with the same number of clusters, such as \textit{RefrigerationDevices} and \textit{CBF}, the execution time variations were primarily attributed to differences in the number of data points.

\begin{table}
\centering
\caption{Speedup (×) of WOA-kMedoids compared to PAM on datasets with more than 300 data points.}
\label{tab05}
\begin{tabular}{lrrl}
\toprule
               Dataset & Number &  k & Speedup \\
\midrule
     EOGVerticalSignal &    362 & 12 &   2.29× \\
  RefrigerationDevices &    375 &  3 &   1.02× \\
          EthanolLevel &    500 &  4 &   1.22× \\
     SemgHandGenderCh2 &    600 &  2 &   1.63× \\
                   CBF &    900 &  3 &   1.50× \\
            TwoLeadECG &   1139 &  2 &   1.24× \\
            MoteStrain &   1252 &  2 &   1.10× \\
               FaceAll &   1690 & 14 &   1.94× \\
   InsectWingbeatSound &   1980 & 11 &   1.77× \\
                  Yoga &   3000 &  2 &   2.11× \\
UWaveGestureLibraryAll &   3582 &  8 &   1.85× \\
           TwoPatterns &   4000 &  4 &   1.85× \\
               ECG5000 &   4500 &  5 &   2.18× \\
       ElectricDevices &   7711 &  7 &   1.89× \\
       StarLightCurves &   8236 &  3 &   1.77× \\
               Average &        &    &   1.69× \\
\bottomrule
\end{tabular}
\end{table}

The Wilcoxon signed-rank test yielded a test statistic of $W = 67.0$ and $p$-value of 0.009 (see Table \ref{tab06}). The test statistic $W$ represents the smaller sum of signed ranks, corresponding to the 10 datasets where PAM was faster, compared to the 15 datasets where WOA-kMedoids was faster. Under the null hypothesis of no performance difference, we would expect roughly equal rank sums ($\approx$162.5 each). The observed $W = 67.0$ deviates substantially from this expectation, with the complementary sum being 258.0, indicating that WOA-kMedoids not only won on more datasets, but also achieved larger performance gains. The $p$-value of 0.009 confirms that this directional bias is statistically significant, supporting our conclusion that WOA-kMedoids demonstrates superior computational efficiency across the dataset collection.

\begin{table}[htbp]
\centering
\caption{Result of the Wilcoxon signed-rank test.}
\label{tab06}
\begin{tabular}{lll}
\toprule
Statistic ($W$) & $p$-value \\
\midrule
67.0 & 0.009 \\
\bottomrule
\end{tabular}
\end{table}

\begin{figure}[htbp]
\centering
\includegraphics[width=\linewidth]{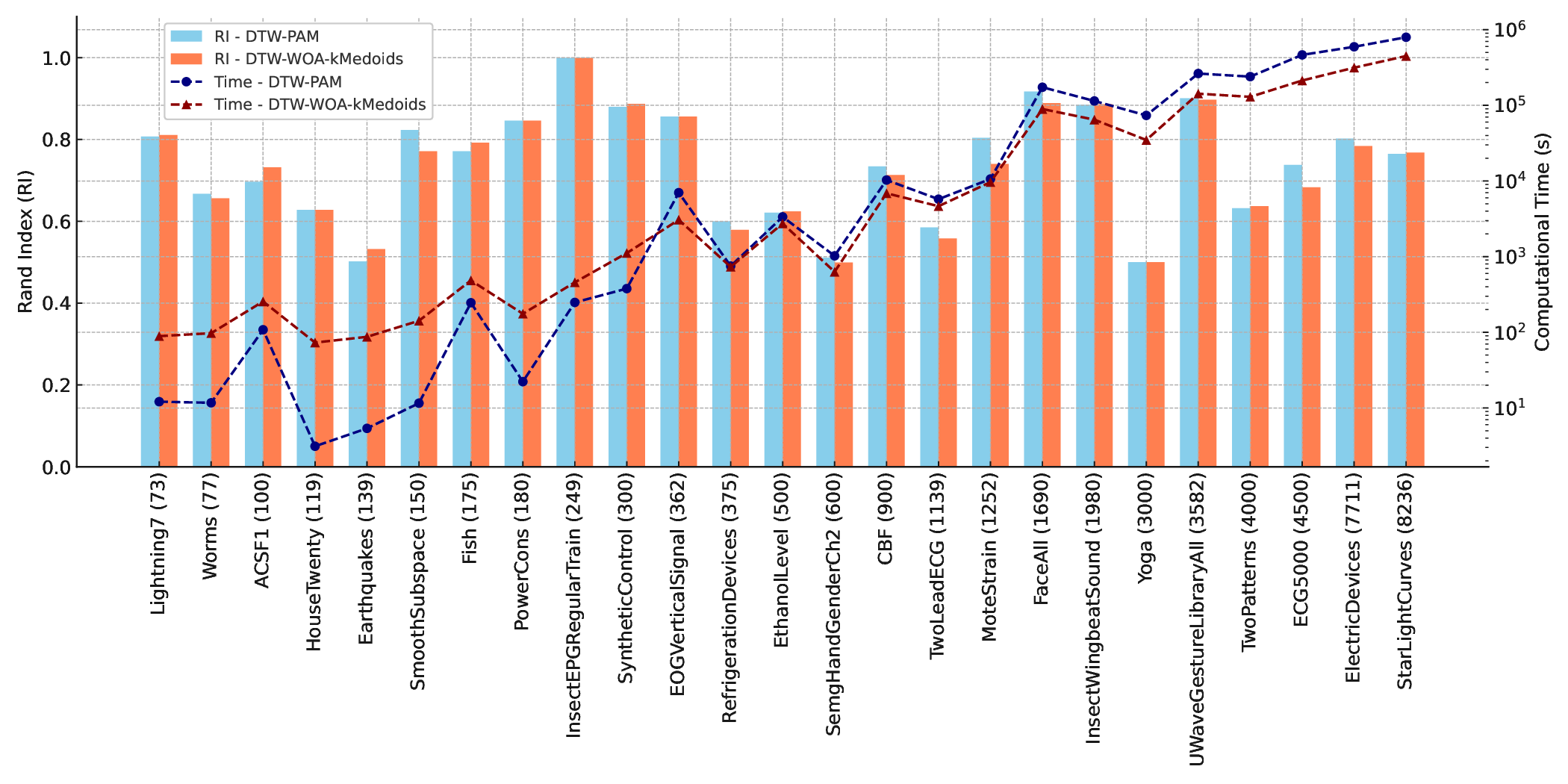}
\caption{Comparison of RI values and computational time (log scale) between DTW-PAM and DTW-WOA-kMedoids across datasets. The bar charts represent the RI values, and the line plots represent the computational time. The numbers in parentheses on the $x$-axis indicate the number of instances in each dataset.}
\label{fig_7}
\end{figure}

Figure \ref{fig_7} summarizes the clustering precision and computational time for DTW-PAM and DTW-WOA-kMedoids across all the 25 datasets.

The results show that WOA-kMedoids maintains comparable accuracy to PAM, while achieving significantly better computational efficiency as the dataset size increases. With datasets arranged in order of increasing size, the WOA-kMedoids' runtime advantage becomes more pronounced for larger datasets, demonstrating superior scalability.

These findings show that WOA-kMedoids strikes an effective balance between clustering accuracy and computational cost, maintaining precision while offering improved efficiency, particularly for larger datasets.

\subsection{Convergence and Diversity Analysis}
To examine how WOA-kMedoids achieves computational efficiency while maintaining clustering quality, we analyzed the optimization behavior of the algorithm using the EOGVerticalSignal dataset as a case study.

\begin{figure}[htbp]
    \centering
    \includegraphics[width=\textwidth]{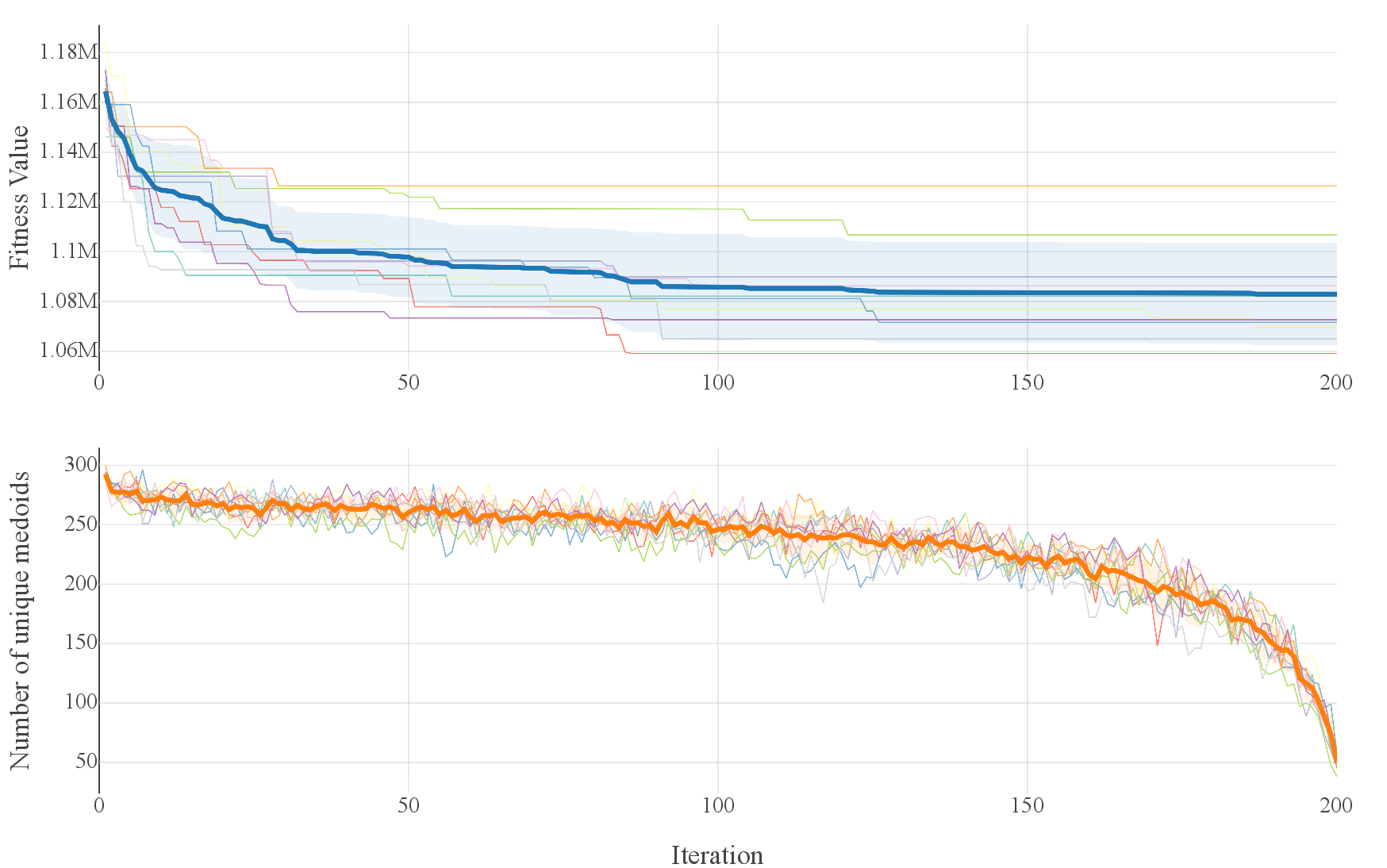}
    \caption{Convergence and diversity evolution of WOA-kMedoids on EOGVerticalSignal dataset over 200 iterations by 10 independent executions. Top: convergence curve showing fitness values; Bottom: population diversity. Bold lines indicate the average of 10 executions.}
    \label{fig_4}
\end{figure}

Figure \ref{fig_4} shows the distinctive optimization pattern of the algorithm. The convergence shows rapid improvement: 90\% of the total fitness improvement occurs by iteration 64, with over 97\% achieved within 100 iterations, on average. The diversity evolution shows a clear two-phase pattern: a gradual reduction from 100\% to 75-85\% during the first 100 iterations, followed by a sharp decline to approximately 17\% by 200 iterations.

The divergent patterns between the fitness convergence and diversity evolution reveal a key characteristic of the algorithm. At iteration 100, while the fitness nearly converged, the diversity remained high. This occurs because whales have discovered near-optimal solutions, but with the parameter $\vec{a}$ still relatively large (decreasing linearly from 2 to 0), they continue exploring a substantial neighborhood around these quality solutions. Only when $\vec{a}$ approaches zero during iterations 100–200 does the exploration radius dramatically shrink, causing the sharp decline in diversity.

This optimization pattern shows how WOA-kMedoids balances efficiency and quality. The adaptive mechanism of the algorithm allows whales to quickly find high-quality solutions while avoiding premature convergence to local optima during exploration, and then naturally transitions to intensive optimization as $\vec{a}$ decreases. This approach enables WOA-kMedoids to match PAM's accuracy of PAM without exhaustively evaluating all the possible medoid swaps.

However, our analysis also revealed a limitation in the current implementation. The dataset achieves over 97\% of its improvement within 100 iterations, yet the algorithm continues for the full 200 iterations with minimal additional benefits. This highlights a common challenge in metaheuristic design: balancing between ensuring convergence and avoiding unnecessary computations.

\subsection{Parameter Sensitivity Analysis}
Using the EOGVerticalSignal dataset, we conducted a sensitivity analysis across thirty-five parameter combinations to evaluate both clustering accuracy (RI) and convergence quality (fitness value). The RI value by ED-PAM and DTW-PAM was 0.856.

\begin{figure}[htbp]
    \centering
    \includegraphics[width=\textwidth]{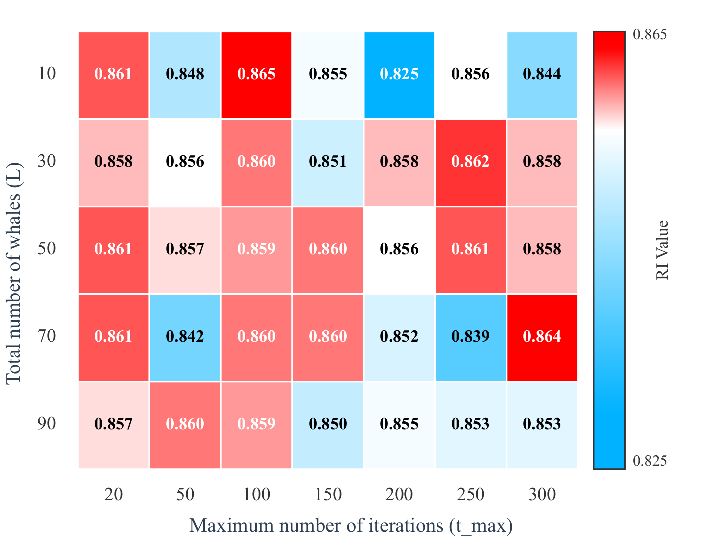}
    \caption{Parameter sensitivity analysis for clustering accuracy (Rand Index). The heatmap shows the RI values across different population sizes ($L$) and iteration counts ($t_{\text{max}}$) for the EOGVerticalSignal dataset. Higher values (red colors) indicate a better clustering performance than PAM. The RI value of ED-PAM and DTW-PAM are 0.856.}
    \label{fig_5}
\end{figure}

Figure~\ref{fig_5} shows stable clustering accuracy, with RI values ranging from 0.825 to 0.865. The highest RI of 0.865 occurred at $L=10, t_{\text{max}}=100$, and most configurations achieved $\text{RI} \geq 0.840$. Notably, the majority of parameter configurations achieved RI values equal to or higher than PAM's result (0.856), demonstrating comparable or superior  clustering performance. 

\begin{figure}[htbp]
    \centering
    \includegraphics[width=\textwidth]{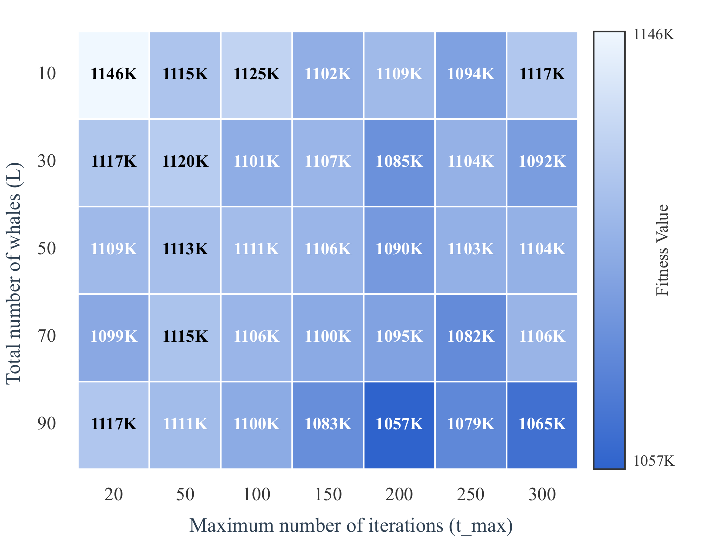}
    \caption{Parameter sensitivity analysis for convergence quality (fitness values). The heatmap displays the fitness values across different population sizes ($L$) and iteration counts ($t_{\text{max}}$) on the EOGVerticalSignal dataset. Lower values (darker blue colors) indicate better convergence.}
    \label{fig_6}
\end{figure}

By contrast, Figure~\ref{fig_6} shows a significant fitness variation between 1,057,146 and 1,146,281 (an 8.4\% range), with convergence consistently improving at higher iteration counts. The lowest fitness value of 1,057,146 is achieved at $L=90, t_{\text{max}}=200$.

This contrast reveals a key evaluation challenge: configurations that perform similarly in clustering can exhibit markedly different optimization states. For example, when $L=10, t_{\text{max}}=20$ and $L=90, t_{\text{max}}=200$ achieved similar RI values (0.861 vs. 0.855), their fitness values differed by 89,135 units, indicating vastly different levels of convergence quality.

The analysis revealed the robustness of the algorithm and its optimization limitations. WOA-kMedoids showed consistent clustering performance across diverse parameter settings, demonstrating tolerance to parameter misspecifications. However, the configuration with the best fitness value ($L=90, t_{\text{max}}=200$) yields a lower RI than the configuration that optimizes RI ($L=10, t_{\text{max}}=100$), suggesting the presence of local optima that affect the cluster quality.

\section{Discussion}
\label{sec:discussion}

\subsection{Algorithmic Advantages }
\label{sec:advantage}

Our comprehensive evaluation using predefined parameters ($L=50$, $t_{\max}=200$) across 25 UCR datasets demonstrates that WOA-kMedoids achieves a favorable algorithmic trade-off, addressing fundamental scalability limitations in traditional clustering approaches. The results revealed three distinct advantages of establishing WOA-kMedoids as a scientifically robust solution for large-scale unsupervised clustering.

The core algorithmic contribution transforms the medoid selection problem from an exhaustive combinatorial search to a guided metaheuristic optimization. The traditional PAM exhibits $O(kn^2)$ complexity because it must evaluate all possible medoid-non-medoid swaps, creating a computational bottleneck that scales quadratically with the dataset size. Our WOA-kMedoids approach reduces this to $O(Lt_{\max}kn)$ complexity, where the linear relationship with $n$ becomes the dominant factor for large datasets.

Empirically, this theoretical advantage manifests as a consistent 1.7$\times$ average speedup on datasets exceeding 300 observations, with a maximum gain of 2.3$\times$ on the EOGVerticalSignal dataset ($n=362$, $k=12$). The Wilcoxon signed-rank test confirmed that this performance advantage was statistically significant ($W=67.0$, $p=0.009$), with WOA-kMedoids demonstrating superior performance in cases where its improvement margin substantially exceeded instances in which PAM performed better. 

Crucially, this computational efficiency is achieved without sacrificing clustering quality. The average Rand Index difference between DTW-WOA-kMedoids (0.731) and DTW-PAM (0.739) was statistically non-significant (Conover test, $p=0.230$), demonstrating that metaheuristic optimization successfully identifies medoid combinations that approximate the globally optimal solutions found through an exhaustive search.

Beyond computational efficiency, WOA-kMedoids exhibited superior algorithmic stability across heterogeneous datasets. The variance in RI values (0.019) is similar to that of both DTW-PAM (0.020) and ED-PAM (0.020), indicating consistent performance across datasets with varying dimensionalities (15-2000), sample sizes (73-8236), and cluster structures ($k=2-14$). This stability was achieved with pre-defined parameters without any optimizations.

The WOA-kMedoids framework demonstrated distance-metric agnostic performance. While our evaluation focused on DTW for time-series data, the design of the algorithm flexibly accommodates various distance measures (Euclidean, Manhattan, cosine similarity, etc.) without requiring modifications to its core optimization strategy. This versatility positions WOA-kMedoids as a general-purpose clustering tool suitable for diverse data types and application domains.

The consistent outperformance of both DTW-based methods over ED-PAM (Friedman test, $p < 0.001$) validates the importance of selecting appropriate distance metrics for time-series clustering, while demonstrating that WOA-kMedoids preserves this advantage while increasing computational efficiency. 

\subsection{Limitations }
\label{sec:limitations}

Although our approach outperforms the conventional PAM in terms of computational cost, our analysis reveals two key shortcomings. First, Figure \ref{fig_3}(a) shows a lower computational speed for smaller datasets with fewer than 300 data points. Second, the convergence analysis (Section 5.4) demonstrates that the representative dataset reaches over 97\% of its improvement within 100 iterations, yet continues running for the full 200 iterations with minimal additional benefit. These  results suggest that our fixed-parameter configuration was suboptimal across different dataset sizes and characteristics.

These limitations highlight the need for adaptive parameter control in WOA-kMedoids. Our sensitivity analysis shows that while the algorithm maintains robust performance across parameter variations (RI range: 0.825-0.865), the configurations that yield optimal clustering do not always align with the optimal convergence states. This suggests that an effective parameter selection must balance multiple competing objectives. Future studies should develop mechanisms to dynamically adjust the population size and termination criteria based on the dataset characteristics, convergence patterns, and solution quality metrics. For example, a multi-objective adaptive control system may optimize the trade-off between the clustering accuracy and computational efficiency. Additionally, the population size would be dynamically scaled based on dataset complexity instead of using fixed parameters across all scenarios.

The choice of distance metric significantly affects the clustering performance. For time-series data clustering, our results confirm that Dynamic Time Warping (DTW) achieves higher accuracy than Euclidean Distance. However, the computational cost of DTW remains a significant concern in our experiments. Future work will explore optimized DTW-based algorithmssuch as FastDTW \cite{salvador2007toward} and IncDTW \cite{leodolter2021incdtw}, which have proven effective in accelerating DTW calculations. Integrating these algorithms with WOA-kMedoids could further enhance their end-to-end effectiveness for time-series clustering.

This framework can be extended by incorporating metaheuristic algorithms beyond the WOA. Although we chose the WOA for its simplicity and strong convergence properties, other swarm-based or evolutionary approaches could effectively optimize medoid selection in k-medoids clustering. Previous research has explored hybrid frameworks that combine WOA with other metaheuristic algorithms \cite{eid2021binary}. Building on these findings, future studies could investigate hybrid optimization strategies that combine the WOA with complementary techniques to improve adaptability and performance across different data scenarios. Examining both alternative and hybrid approaches within a unified experimental framework will deepen our understanding of metaheuristic-driven clustering methods and advance the development of more generalizable solutions.

\section{Conclusions}
\label{sec:conclusion}

We propose WOA-kMedoids, a novel unsupervised clustering algorithm that combines the Whale Optimization Algorithm with k-medoids clustering. Through population-based optimization of medoid selection, our method delivers significant computational efficiency gains compared with traditional PAM, especially for large datasets, while maintaining comparable clustering accuracy.

Our comprehensive evaluation across 25 UCR time-series datasets shows that WOA-kMedoids reduces the computational complexity from $O(kn^2)$ to $O(Lt_{\max}kn)$. This achieves an average 1.7$\times$ speedup on datasets exceeding 300 observations without sacrificing clustering quality (average RI: 0.731 vs. PAM's 0.739, statistically non-significant difference). The adaptive exploration-exploitation balance of the algorithm enables rapid convergence while maintaining stability across heterogeneous datasets.

Currently, we use a predefined parameter setting, which leads to slower performance on small datasets ($<$300 observations) and continues iterations beyond convergence, indicating the need for adaptive parameter-control mechanisms. Future studies should investigate dynamic parameter adjustment and hybrid metaheuristic approaches to enhance versatility and robustness.

WOA-kMedoids offers a scalable and generalizable solution for large-scale clustering applications. This is particularly valuable for domains where traditional k-medoids methods are computationally infeasible, such as IoT anomaly detection, biomedical signal analysis, and customer behavior clustering. The distance-metric flexibility and linear scalability of the method make it an effective tool for knowledge discovery in extensive unlabeled datasets across various real-world domains.

The R code for WOA-kMedoids is Publicly available at: https://CRAN.R-project.org/package=WOAkMedoids

\bibliographystyle{unsrt}  
\bibliography{reference}

\end{document}